\documentclass[letterpaper]{article} 
\usepackage{aaai2026}  
\usepackage{times}  
\usepackage{helvet}  
\usepackage{courier}  
\usepackage[hyphens]{url}  
\usepackage{graphicx} 
\usepackage{natbib}  
\usepackage{caption} 
\usepackage{algorithm}
\usepackage{multirow, makecell}
\usepackage{colortbl}
\usepackage{algpseudocode}
\usepackage{xspace}
\usepackage{multicol}
\usepackage{tikz}
\usepackage{array}
\usepackage{xcolor}
\usepackage{subfigure}
\usepackage{soul}

\usepackage{amsmath}

\newcommand{\City}{\texttt{City}\xspace}

\newcommand{\SBengaluru}{\texttt{A}\xspace}
\newcommand{\SChennai}{\texttt{B}\xspace}
\newcommand{\SDelhi}{\texttt{C}\xspace}
\newcommand{\SKolkata}{\texttt{D}\xspace}
\newcommand{\SMumbai}{\texttt{E}\xspace}

\newcommand{\Bengaluru}{\texttt{City A}\xspace}
\newcommand{\Chennai}{\texttt{City B}\xspace}
\newcommand{\Delhi}{\texttt{City C}\xspace}
\newcommand{\Kolkata}{\texttt{City D}\xspace}
\newcommand{\Mumbai}{\texttt{City E}\xspace}

\newcommand{\ALGO}{\texttt{LaMSUM}\xspace}
\renewcommand{\SS}{\ensuremath{\mathcal S}\xspace}
\newcommand{\TT}{\ensuremath{\mathcal T}\xspace}

\makeatletter
\newcommand{\thickhline}{%
    \noalign {\ifnum 0=`}\fi \hrule height 1.1pt
    \futurelet \reserved@a \@xhline
}
\newcolumntype{"}{@{\hskip\tabcolsep\vrule width 1pt\hskip\tabcolsep}}
\makeatother

\newcolumntype{?}{!{\vrule width 1.1pt}}

\newcommand*\circled[1]{\tikz[baseline=(char.base)]{
            \node[shape=circle,draw,inner sep=2pt] (char) {#1};}}

\definecolor{Gray}{gray}{0.92}
\definecolor{lightgray}{gray}{0.95}
\definecolor{darkgray}{gray}{0.85}

\newcommand{\answerYes}[1]{\textcolor{blue}{#1}} 
 
\newcommand{\answerNA}[1]{\textcolor{gray}{#1}}

\makeatletter
\def\blfootnote{\xdef\@thefnmark{}\@footnotetext}
\makeatother

\title{LaMSUM: Amplifying Voices Against Harassment through LLM Guided Extractive Summarization of User Incident Reports}

\author{
   Garima Chhikara\textsuperscript{\rm 1,\rm2},
   Anurag Sharma\textsuperscript{\rm 3},
   V. Gurucharan\textsuperscript{\rm 4} \\
   Kripabandhu Ghosh\textsuperscript{\rm 5}, 
   Abhijnan Chakraborty\textsuperscript{\rm 3}
}
\affiliations {
   \textsuperscript{\rm 1} Indian Institute of Technology Delhi, India \\
   \textsuperscript{\rm 2} Delhi Technological University, India \\
   \textsuperscript{\rm 3} Indian Institute of Technology Kharagpur, India \\
   \textsuperscript{\rm 4} Collaborative Dynamics, Texas, USA \\
   \textsuperscript{\rm 5} Indian Institute of Science Education and Research Kolkata, India \\
}

\begin{document}

\maketitle

\begin{abstract}
Citizen reporting platforms help the public and authorities stay informed about sexual harassment incidents.  
However, the high volume of data shared on these platforms makes reviewing each individual case challenging. Therefore, a summarization algorithm capable of processing and understanding various code-mixed languages is essential. 
In recent years, Large Language Models (LLMs) have shown exceptional performance in NLP tasks, including summarization. LLMs inherently produce abstractive summaries by paraphrasing the original text, while the generation of extractive summaries -- selecting specific subsets from the original text -- through LLMs remains largely unexplored. Moreover, LLMs have a limited context window size, restricting the amount of data that can be processed at once. We tackle these challenges by introducing \ALGO, a novel multi-level framework combining summarization with different voting methods to generate extractive summaries for \textit{large} collections of incident reports using LLMs. 
Extensive evaluation using four popular LLMs (Llama, Mistral, Claude and GPT-4o) demonstrates that \ALGO outperforms state-of-the-art extractive summarization methods. Overall, this work represents one of the first attempts to achieve extractive summarization through LLMs,
and is likely to support stakeholders by offering a comprehensive overview and enabling them to develop effective policies to minimize incidents of unwarranted harassment.
\\
\textit{{\color{red} Warning:} This paper contains content that may be disturbing or upsetting.}
\end{abstract}

%

\section{Introduction}

\begin{table*}[t]
\small
\centering
\begin{tabular}{p{1.8cm} | p{15cm}}
\thickhline
\rowcolor{Gray} \textbf{Category} & \textbf{Post} \\ \thickhline
\multirow{1}{*}{Robbery} & This incident took place in the evening.Two bikers came on a bike and snatched Rs.17000 from an old lady at gun point. \\ \hline
\multirow{2}{*}{Stalking} & I was stalked by a guy who followed me for days and also he sent me letters on my doorstep saying he was madly in love with me and he is obsessed with my body. \\ \hline
\multirow{3}{*}{Sexual Invites} & I was walking on footpath to a place nearby to meet my friends. A man was driving his car on parallel road and was continuously passing vulgar comments. I ignored but 15 min later he stopped his car and said \textit{``chalri h ky, paise h mere pas''} (I have money, you want to come with me?). I screamed and asked for help from people around me. \\ \hline
Mas*****tion in public & One day I reached my school prior to the school timings mistakenly. One of the drivers called me and my friend and started mas*****ting in front of us. \\ \hline
\multirow{3}{*}{Ogling} & I was going to my coaching by driving my scooty when suddenly some boys came on a bike. They started making cheap comments and teasing me. It was late and the area was secluded. I got scared as they started revolving their bike around me. I started driving in the direction of a crowded place that is when they left. \\ \hline
Showing Po**ography & A man in a car parked outside was watching po** when I saw him. He turned his device towards me and did offensive hand gestures inviting me. \\ \hline
\multirow{2}{*}{Sexual Assault} & When I was 7 years old, the shopkeeper removed my clothes and started touching me everywhere.
He also tried to do it to another girl and failed. \\ \hline
Domestic \hspace{0.5cm} Violence & My husband always doubts on my character and doesn't allow me to go outside alone, uses very vulgar language and beats me. \\
\thickhline
\end{tabular}
\caption{Examples of harassment cases shared on an incident reporting platform. Proactive action by authorities and citizens can help prevent numerous such incidents. Providing stakeholders with a concise overview of incidents occurring in a specific area is crucial and this can be effectively achieved by utilizing summarization algorithms.}
\label{tab:incident}
\end{table*}

In recent decades, the widespread availability of the internet has provided seamless access to online platforms to millions of people.
Governments worldwide are increasingly utilizing these platforms to gather information directly from citizens -- referred to as \textit{Citizen Reporting}~\cite{kopackova2019citizen}.
By leveraging tools such as mobile applications, web-based portals, and social media integrations, citizen reporting platforms establish a direct and efficient communication link between individuals and the relevant authorities, enabling faster issue resolution and facilitating active public participation in community improvement. 
Beyond immediate problem solving, real-time data gathered through these platforms contributes valuable information for urban planning and proactive measures, paving the way for more efficient and adaptive communities.
Citizen reporting typically addresses topics such as community issues, environmental challenges, crime prevention, public health, and disaster response~\cite{shin2024systematic}. 

A special category of citizen reporting platforms, such as \textit{Safe City}~(\url{https://webapp.safecity.in}), 
\textit{SHe-Box}~({\url{https://shebox.wcd.gov.in}}), and \textit{JDoe}~({\url{https://jdoe.io}}), allow people to post incidents of sexual harassment, domestic abuse, violence and assault.
Table~\ref{tab:incident} showcases some example incidents 
shared by users on one such platform. 
While such horrific incidents cannot be entirely avoided through reporting alone, these \textit{incident reporting platforms} 
can play a crucial role in preventing certain cases of sexual assault. By enabling users to analyze reported incidents, assess the safety of specific locations, and make informed decisions when traveling to potential hotspots, these platforms contribute to enhanced personal safety and awareness.
Similarly, the local authorities 
can also benefit from these platforms to assess emerging cases, identify the underlying factors and determine proactive measures for effective resolution. 
However, the challenge for the authorities is to navigate the high volume of information in such platforms. 
Manually reviewing all posts is often impractical, {\it necessitating a summarization algorithm that can identify and select posts that are diverse as well as representative of the original data}. 
Additionally, incident reporting platforms often feature a curated selection of posts on their homepage to showcase their core purpose, mission, and services. {\it This deliberate selection also acts as a form of summarization}.

Summarization algorithms are of two types: 
`extractive' and `abstractive'. In {\it extractive summarization}, the 
algorithm selects a subset 
representative of the original text \cite{xu-etal-2020-discourse, zhong-etal-2020-extractive, zhang-etal-2022-hegel, dash2018fairsumm, zhang-etal-2023-diffusum}. 
In contrast, {\it abstractive summarization} algorithms generate 
summaries that capture the essence of the original text, often paraphrasing the content~\cite{pu2023summarization}. 
For incident reporting platforms, extractive summarization is more suitable, as the goal is not to paraphrase the posts but to select a few that accurately capture a snapshot of the original content. {\it When summarizing such sensitive posts, preserving the user’s exact words is essential, making extractive summarization particularly valuable in maintaining authenticity and context}.

Several extractive summarization algorithms for user generated content have been proposed in the literature, primarily for text written in English~\cite{10.1145/3462757.3466092, 10.1145/3477314.3507256, 10.1145/3397271.3401269, jia-etal-2020-neural}. But there are several countries where English is not the primary language and users frequently communicate in code-mixed forms. For instance, India recognizes 22 official languages and users often post in Hinglish (a mix of Hindi and English). 
Such multilinguality limits the applicability of existing algorithms for extractive summarization of incident posts.

In recent years, Large Language Models (LLMs) have 
demonstrated 
very good performance across various tasks in multilingual and code-mixed settings~\cite{rlhf1, NEURIPS2020_1457c0d6, Tang2023-jl, jin2024comprehensivesurvey}. Plus, summaries generated by LLMs showcase high coherence and are overwhelmingly 
preferred by human evaluators over other baseline algorithms~\cite{pu2023summarization, liu2023learning}. These prior results motivated us to investigate the utility of LLMs for extractive summarization of large volumes of user generated posts. 
However, we encountered two significant limitations which hinder the immediate application of LLMs for extractive summarization:
\begin{enumerate}
    \item As generative models, LLMs 
perform abstractive summarization by paraphrasing rather than selecting 
the most relevant sentences (as shown in Figure~\ref{fig:current_state}) \cite{spectrum}.
\item Due to the finite size of the context window, LLMs cannot handle long texts in a single input, underscoring the need for a method that allows for processing long text \cite{jin2024llmmaybelonglmselfextend}.
\end{enumerate}

To overcome these limitations, in this paper, we present a novel framework \ALGO (\textbf{La}rge Language \textbf{M}odel based Extractive \textbf{SUM}marization) that integrates LLM-generated summaries with voting algorithms borrowed from Social Choice Theory \cite{brandt2016handbook}. 
Our judicious application of voting algorithms with a multi-level summarization framework ensures that \ALGO outperforms the state-of-the-art fine-tuned summarization models. 
In summary, in this work, we make the following contributions:
\begin{itemize}
    \item We propose a novel framework \ALGO which can produce 
    extractive summaries from large (having $>$30K tokens) collection of user generated content. 
    \ALGO considers a multi-level summarization model that utilizes 
    voting algorithms to combine LLM outputs 
    to generate robust summaries.
    \item Extensive experiments with incident posts demonstrate that \ALGO outperforms the state-of-the-art extractive summarization algorithms.
\end{itemize}


To our knowledge, this is the first work to implement extractive summarization of a large collection of user-generated texts using LLMs by combining summarization with voting algorithms. At the same time, we demonstrate the effectiveness of such algorithms to facilitate data-driven decision-making promoting safer communities by providing actionable insights into reported incidents. 
Code is available at https://github.com/garimachhikara128/LaMSUM

\section{Background and Related Work}
\label{sec:related_work}
In this section, we review the relevant prior works that provide the foundation for our current research.

\begin{figure}[t]
\centering
\includegraphics[width=0.97\linewidth,trim={170 45 220 25},clip]{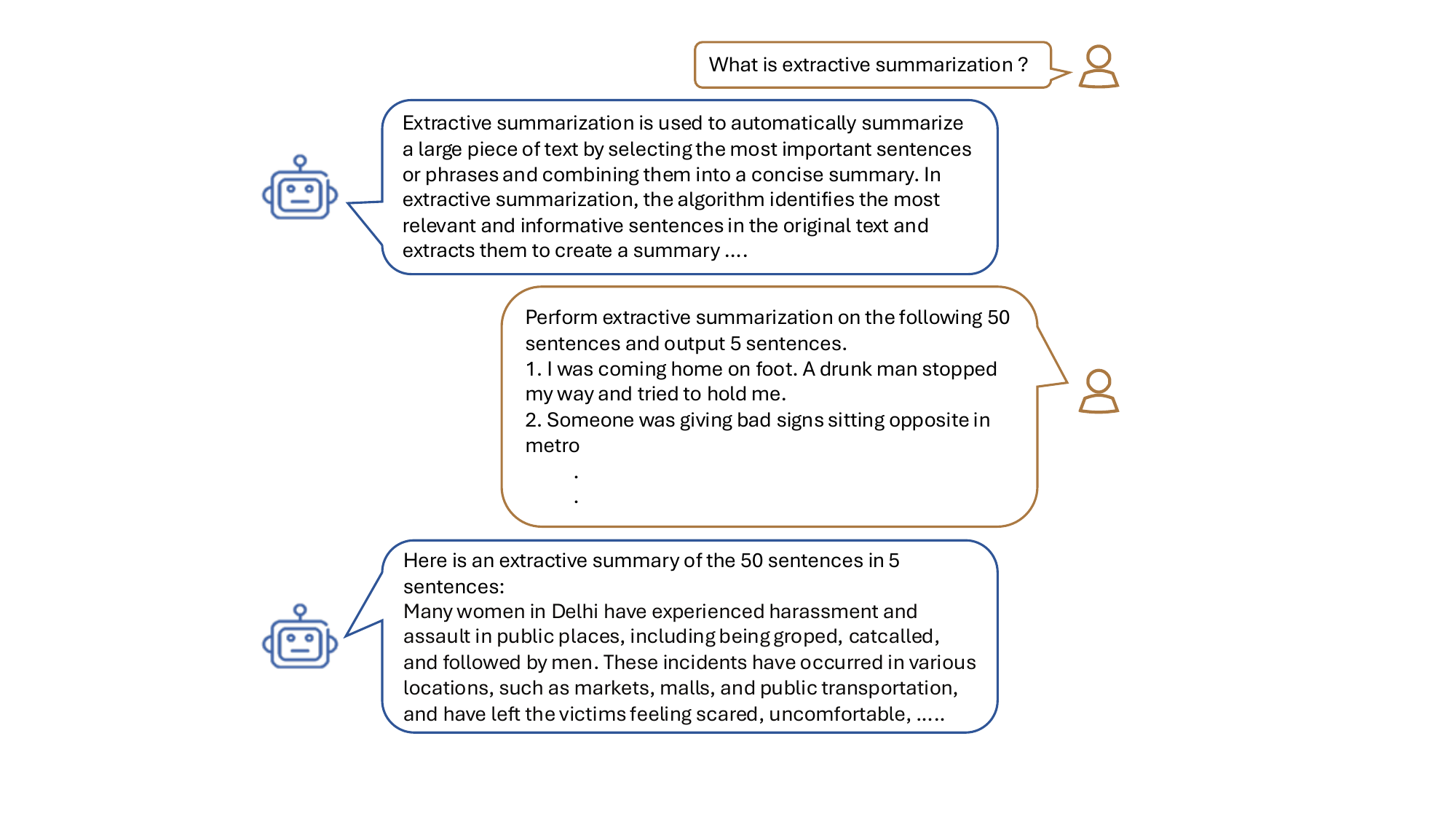}
\caption{Current LLMs, by default, produce abstractive summaries. \texttt{Llama-3.3-70b}, despite  specifically prompted for extractive summarization, generates abstractive summaries. 
This behavior underscores the need for a targeted approach to enable LLMs to effectively generate extractive summaries.}
\label{fig:current_state}
\vspace{-1mm}
\end{figure}

\vspace{1mm}
\noindent \textbf{AI Solution through Citizen Reporting} \\
Web and social media platforms receive posts on sensitive issues such as online harassment, hate speech, abusive behavior, violence etc. 
Abuse experienced by users leads to mental stress often forcing them to leave the platform \cite{10.1145/3290605.3300232, 10.1145/3491102.3501879, Kim_Razi}.  
Several AI-powered solutions have been designed to address solutions to these critical issues. 
Machine learning based classifiers and language models are utilized to detect the cases of sexual abuse, hate speech, offensive language, human trafficking and harassment cases \cite{sawhney-etal-2021-multitask, Hassan_Poudel, davidson, singh2025rethinking,  Upadhayay2021CombatingHT, stoop-etal-2019-detecting, ghosh-chowdhury-etal-2019-speak}. 
Modelling the cyberbullying behavior using social network and language-based features can improve the classifier performance \cite{Ziems_Vigfusson_Morstatter_2020, Olteanu_Castillo_Boy_Varshney_2018}.
Development of a mobile computing based reporting tool empowers individuals with intellectual and developmental disabilities (I/DD) to self report abuse and share the incident with the intended group \cite{10.1145/3411764.3445150, 10.1145/3411764.3445154}.
With the rise of public figures encouraging women to speak up, the number of non-anonymous self-reported assault stories has increased \cite{ElSherief_Belding_Nguyen_2017}. 
Counterspeech is proven to be a viable alternative to blocking or suspending problematic messages or accounts, as it better aligns with the principles of free speech \cite{MathewSaha}.
Conversational agents (CAs) have attracted significant interest as potential counselors due to their features, such as anonymity, which can help address many challenges associated with human-human interaction~\cite{10.1145/3411764.3445133}.

\vspace{1mm}
\noindent \textbf{Large Language Models (LLMs) for Summarization} \\
LLMs are now being extensively used for summarization~\cite{NEURIPS2020_1457c0d6, Tang2023-jl, jin2024comprehensivesurvey}. 
Multiple works have proposed few-shot learning frameworks for the abstractive summarization of news, documents, webpages, and generic texts~\cite{zhang-etal-2023, tang-etal-2023-context, yang2023exploring, brazinskas-etal-2020-shot, laskar-etal-2023-systematic}, but their primary focus remains on short documents that can fit in the LLM context window. Researchers have also observed that human evaluators are increasingly preferring LLM-generated summaries compared to other baselines~\cite{10.1162/tacl_a_00632, wu-etal-2024-less, goyal2023news, zhang-etal-2023-summit, liu2023learning}. Despite the advancements, recent studies have 
also uncovered factual inaccuracies and inconsistencies in LLM-generated summaries~\cite{tang2024tofueval, tam-etal-2023-evaluating, luo2023chatgpt, laban-etal-2023-summedits}.

\begin{figure*}[t]
\centering
\includegraphics[scale=0.38,trim={0 0 0 5},clip]{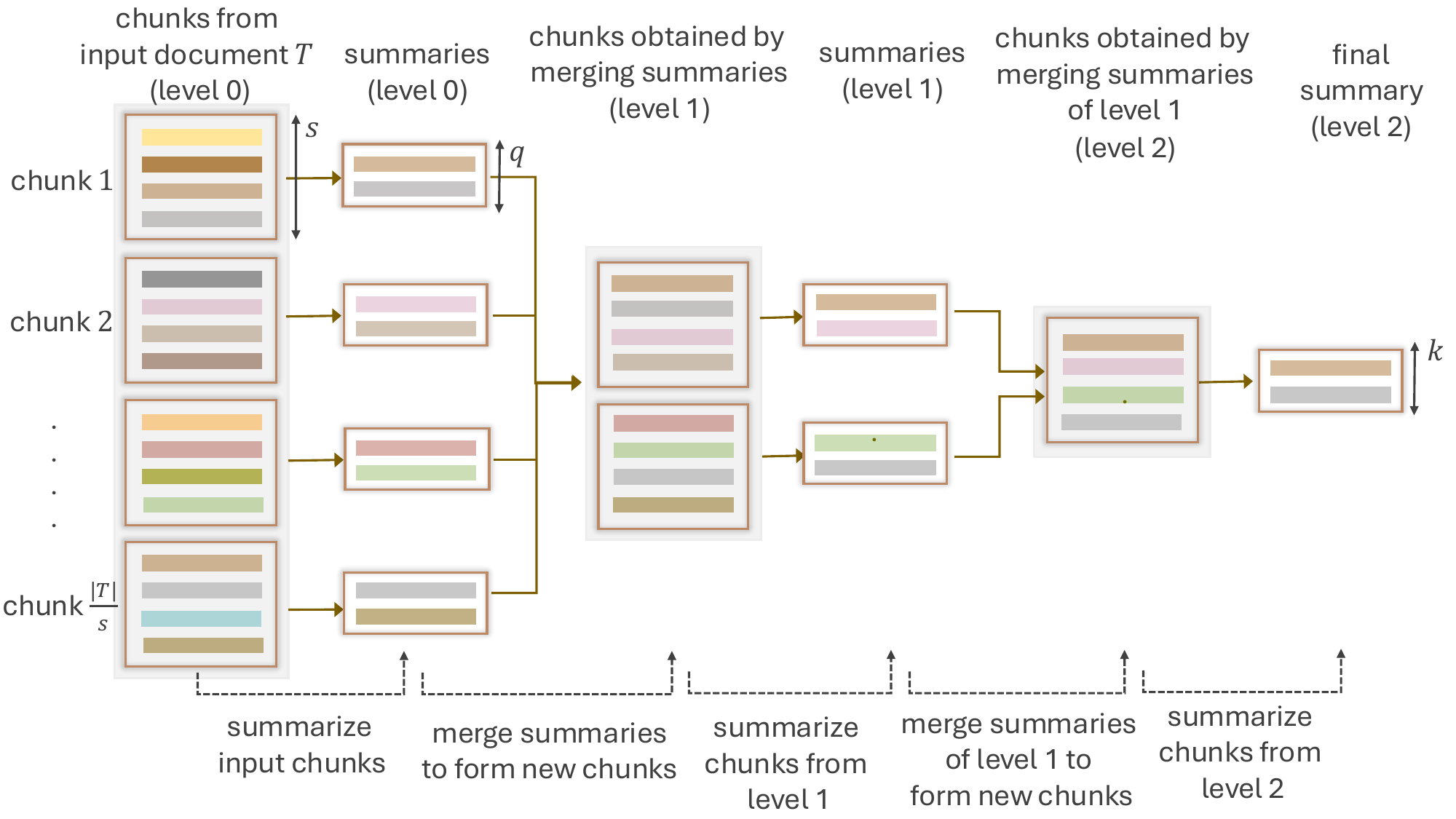}
\caption{\ALGO: Multi-level framework for extractive summarization of large user-generated text. Input set $\TT$ (level 0) is divided into $\lceil{\frac{|\TT|}{s}}\rceil$ chunks each of size s. From each chunk a summary is produced of size $q$ (refer Figure \ref{fig:shuffle}), $q$ length summaries from $\lceil{\frac{|\TT|}{s}}\rceil$ chunks are merged to form the input for the next level i.e., level 1. Iteratively the same procedure is repeated till we obtain a summary of size $k$.
We set $q=k$ to ensure our algorithm can effectively handle the worst-case scenario where all the textual units in the final summary may come from the same input chunk.} 
\label{fig:hierarchy}
\end{figure*}

\vspace{1mm}
\noindent \textbf{Extractive Summarization through LLMs: The Current State} \\
By default, LLMs produce abstractive summaries, meaning that the summary text is distinct from the input text, even when it is instructed to do otherwise. To illustrate this, we present a small example in Figure~\ref{fig:current_state}. An LLM, when prompted, could clearly explain extractive summarization, 
yet, when we instructed it to perform extractive summarization on a set of 50 sentences, it failed to do so and instead generated an abstractive summary. 
Prior to our current work, only two studies attempted to perform similar tasks.
Zhang et al. attempted summarization of short news articles using GPT 3.5 \cite{zhang-etal-2023} , while Chang et al. attempted abstractive summarization for book-length documents \cite{chang2024booookscore}. 
However, both these approaches suffer from practical limitations such as lack of contextual dependencies in user generated text and the problem with positional bias.  

\vspace{1mm}
\noindent
To the best of our knowledge, ours is the first attempt to perform extractive summarization on a large collection of user generated texts through LLMs, while tackling the challenge of positional bias. We describe our proposal in detail in the next section.

\section{\ALGO: Generating Extractive Summaries through LLMs}
In this section, we define the problem statement formally and introduce our novel summarization framework \ALGO (\textbf{La}rge Language \textbf{M}odel based Extractive \textbf{SUM}marization) that leverages LLMs to summarize large user-generated text.

\subsection{Task Formulation}
\label{subsec:task}
Let $\TT = \{t_1, t_2, \ldots t_N\}$ represent a collection of posts, also referred to as a set of textual units.
Our summarization algorithm takes $\TT$ and an integer $k$ as input, where $\TT$ denotes the entire set of textual units and $k$ specifies the desired number of units in the summary. 
The task is to output a summary $\SS \subseteq \TT$ such that $|\SS| = k$. The summary $\SS$ would be evaluated based on its alignment with the preferences of gold standard summarizers.
If the context window size of an LLM is $W$, we assume $\TT$
is too large to fit in a single context window. 

\subsection{Multi-Level Summarization}
\label{sec:multi-level-summarization}
LLMs have a limited context window, making it impossible to input large text collections all at once
within a single window. Consequently, the input must be divided into smaller chunks to perform the desired task~\cite{chang2024booookscore}. 
Thus, \ALGO employs a multi-level framework for extractive summarization, enabling it to consider  input data of any size 
(detailed in Figure \ref{fig:hierarchy}).

The set $\TT$, which contains the original textual units, is provided as input at level 0 and is divided into $\lceil{\frac{|\TT|}{s}}\rceil$ number of chunks of size $s$. From each chunk of size $s$, we generate a summary 
of size $q$ (where $q<s$), and repeat this process for all 
$\lceil{\frac{|\TT|}{s}}\rceil$ chunks.\footnote{Note that a chunk of size $s$ refers to a chunk containing $s$ textual units.
Likewise, a summary of size $q$ indicates a summary of $q$ textual units. $|\TT|$ denotes the number of textual units present in $\TT$.} 
We then merge all these $q$ length summaries obtained from level 0 to form an input for the next level i.e., level 1. We repeatedly perform this process until we obtain the final summary of length $k$. Note that the last chunk may be less than $q$ in size, in such a case we move all the textual units of the respective chunk to the next level (refer Algorithm \ref{algo:summarization}). 
Following a long line of research on extractive summarization~\cite{nallapati2017summarunner, liu-lapata-2019-text}, we have used the summary length as the stopping criterion for our algorithm. Moreover, while evaluating the performance, it is essential for the LLM-generated summary to match the length of the human-written reference summary; otherwise, the ROUGE score would not offer a fair comparison.

An alternative 
strategy would be to divide the input $\TT$ into $\frac{|\TT|}{s}$ chunks each of size $s$ and from each chunk select $\frac{k \cdot s}{|\TT|}$ sentences to be included in the summary. 
However, this approach assumes a uniform distribution of potential candidates across chunks that can be included in the final summary.
In \ALGO, we keep $q=k$ i.e., we extract $k$ 
textual units from each chunk, eliminating the chance of missing any potential candidate.
In the worst-case scenario, all $k$ units in the final summary can come from a single chunk, and our algorithm can handle such cases effectively, as we keep $q=k$. Ablation study for different values of $q$ is discussed in Section \ref{sec:ablation}.

It is important to note that we are dealing with user-generated posts, which 
lack contextual connections. Unlike book summarization, where chapters are interconnected and the context of previous chapters is crucial for summarizing the current one, posts are generally standalone and contextually independent. Thus, our approach of independently deriving summaries from each chunk works well in our setup, as each textual unit operates independently of the others and there are no long-range dependencies. 
However, social media posts can also be contextually connected, including the original post, comments from other users and reposts of others' content. In such cases, our existing framework can be adapted by incorporating clustering at different levels – such as the original post, comments, and comment threads. The \ALGO framework is extendable enough to be further applied at each hierarchical level.

\begin{figure}[t]
\centering
\includegraphics[trim={120 0 80 0},clip, width=0.95\columnwidth]{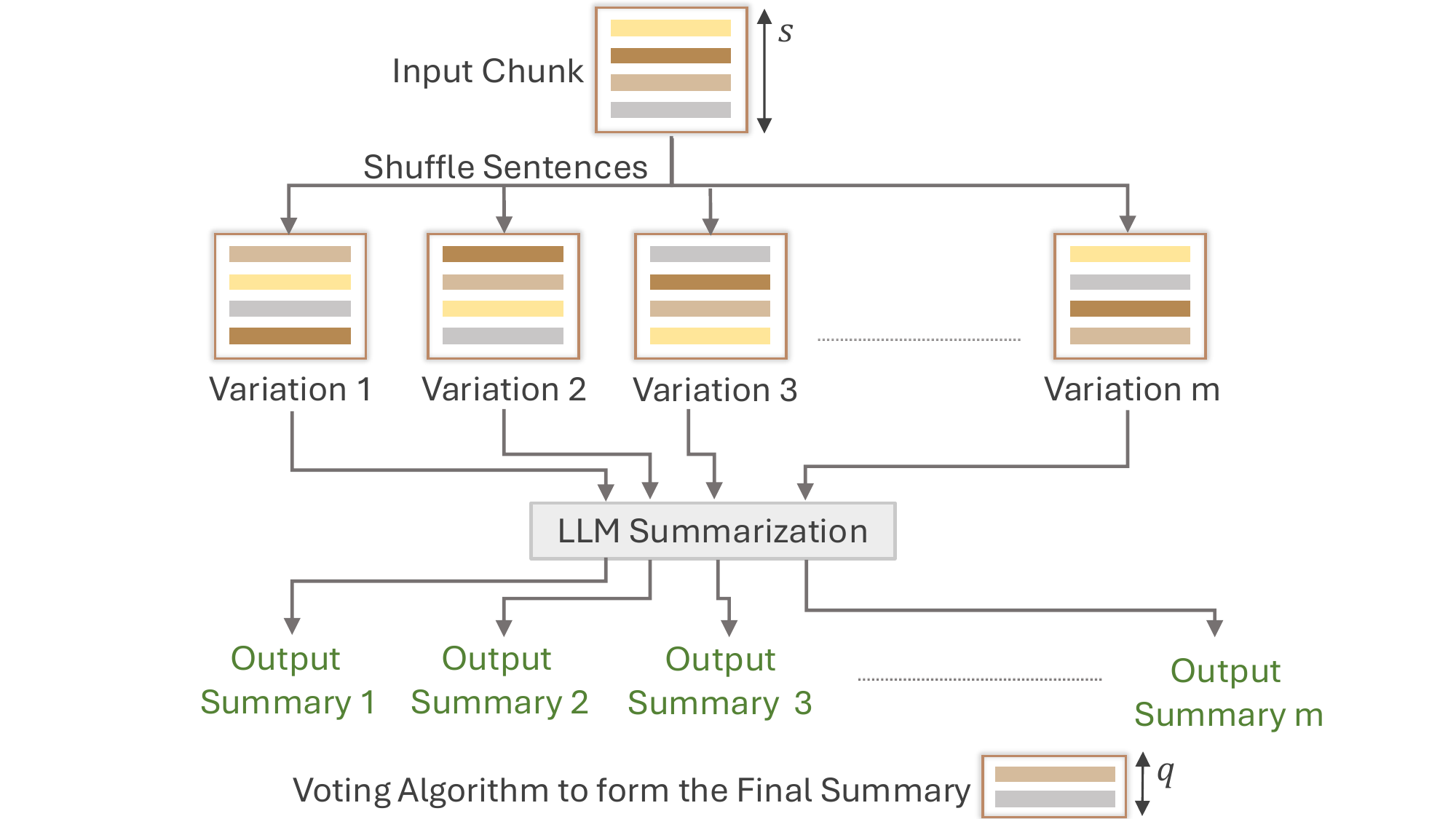}
\caption{Textual units (e.g., posts) in the input chunk are shuffled to account for the positional bias. $m$ different chunk variations are obtained through shuffling, which are subsequently summarized using LLMs. $m$ summaries are then aggregated by voting algorithms to get the final summary.}
\label{fig:shuffle}
\end{figure}

\subsection{Summarizing a Chunk}
\label{subsec:summ_chunk}
Next, we discuss how \ALGO summarizes a chunk (Algorithm \ref{algo:chunk}) by tackling the positional bias in LLMs and leveraging voting algorithms drawn from Social Choice Theory~\cite{brandt2016handbook}.

\subsubsection{Tackling Positional Bias:}
Prior research ~\cite{brown2023how, zhang-etal-2023, jung-etal-2019-earlier, wu-etal-2024-less} has highlighted that summarization using LLM is prone to positional bias, i.e., the sentences located in certain positions, such as the beginning of articles, are more likely to be considered in the summary. To address this issue and generate a robust summary, we create $m$ different variations by shuffling the textual units within the input chunk. This ensures that each unit has the opportunity to appear in different positions within the input text (refer Figure~\ref{fig:shuffle}). 

\subsubsection{Zero-Shot Prompting:}
\label{subsec:zero_shot_prompt}
For each input chunk, we 
obtain $m$ different summaries (one for each variation) by prompting the LLM. We employ the following two prompt strategies to obtain the summaries (detailed in Appendix Table \ref{tab:prompts}):

\noindent \circled{1} Select most suitable units that summarize the input text. \\
\noindent \circled{2} Generate a ranked list in descending order of preference.

\begin{algorithm}[t]
\small
\caption{Algorithm for multi-level summarization}\label{algo:summarization}
\begin{algorithmic}
\State \textbf{Input}: ${\TT, k, s, q, m}$
\State ${\SS = \{ \}}$ \Comment{$\SS$ stores the final summary}
\While{$|\SS| < k$} \Comment{until $k$ length summary is obtained} 
    \State $n_{chunks} = \lceil{\frac{|\TT|}{s}}\rceil$ \Comment{number of chunks in set $\TT$}
    \State $L = \{\}$ \Comment{$L$ stores the results of a given level}
    \For{$i \gets 1$ to $n_{chunks}$}
        \State $si = (i-1) * s$ \Comment{starting index of chunk}
        \State $ei = i * s$ \Comment{ending index of the chunk}
        \If{$i = n_{chunks}$} \Comment{if last chunk}
            \State $ei = |\TT|$ \Comment{ei is equal to length of $\TT$}
        \EndIf
        \State $width = ei - si$ \Comment{no. of textual units in a chunk}

        \If{$width <= q$} \Comment{if last chunk}
            \State $L = L \cup t_{si} \cup t_{si+1} \cup ... \cup t_{ei-1}$ \\ \Comment{add all textual units to the result}
        \Else
            \State $L = L \cup \textproc{ChunkResult}(\TT, si, ei, q, m)$  \\ \Comment{add summary of each chunk to result $L$}
        \EndIf

    \EndFor
    \State $\TT = L$ \Comment{update the input $\TT$ for the next level}
    \State $\SS = \SS \cup L$
\EndWhile
\State \textbf{Output}: $\SS$
\end{algorithmic}
\end{algorithm}

\begin{algorithm}[t]
\small
\caption{Algorithm for summarization of a chunk}\label{algo:chunk}
\begin{algorithmic}
\Function{ChunkResult}{$\TT, si, ei, q, m$}
\State $X = \{\}$
\For{$i \gets 1$ to $m$} \Comment{for each variation of a chunk}
\State $V = \textproc{Shuffle}(\TT,si,ei,i)$ \Comment{shuffle with state $i$}
\State $R = \textproc{LLM}(V,q)$ \Comment{obtain LLM summary}
\State $C = \textproc{Check}(R,\TT,si,ei)$ \Comment{output calibration}
\State $X.add(C)$
\EndFor
\State return $\textproc{VOTING}(X,q)$ \Comment{voting for final summary}
\EndFunction
\end{algorithmic}
\end{algorithm}

\begin{table*}[t]
\footnotesize
\centering
\begin{tabular}{p{10cm} | p{6.7cm}} 
\thickhline
\rowcolor{Gray} \textbf{Original Post} & 
\textbf{LLM Modified Output} \\
\thickhline
In a train some people were staring me continuously. It was very uncomfortable. & 
Some people were staring me continuously in a train.\\
\hline
We were going to metro station, a biker started following us. When we shouted, he rode away. & A biker followed us and rode away when we shouted. \\ \hline
Some boy do dirty comments on me and my religion. & 
Boys made dirty comments about my religion. \\
\thickhline
\end{tabular}
\caption{Examples illustrating that LLMs when selecting textual units for summarization, often demonstrate a propensity to alter certain words or introduce new ones.}
\label{tab:changes}
\end{table*}

\subsubsection{Output Calibration:}
LLMs may 
alter certain words from the input text while generating extractive summaries, as shown in Table~\ref{tab:changes}. Thus, we perform additional checks to ensure that the textual units selected in the summary are indeed a subset of 
$\TT$. If the post selected by the LLM (say $x$) is not present in the original text $\TT$, we identify the post with the closest resemblance to $x$ by computing the edit distance~\cite{682181}. 
LLMs may also hallucinate, generating new sentences rather than selecting units from the input. In such cases, the edit distance between the generated unit $x$ and the original textual units tends to be high. 
To address this, we adopt an alternative approach -- we extract key elements such as nouns, verbs, and adjectives from the newly generated sentence $x$. Then we search for the exact same set of keywords with the textual units in $\TT$ and identify the post with the highest number of matching keywords as the closest match to $x$ (refer Algorithm \ref{algo:calibration} in Appendix).



\subsection{Reimagining Summarization as an Election}
As mentioned earlier, for a given chunk, we obtain $m$ summaries -- one for each variation. {\it We imagine the process of creating the final summary from these $m$ summaries to be a \underline{multi-winner election}}, where the textual units in $m$ summaries correspond to ballots (candidates) and the role of the voting algorithm is to pick $q$ winners. 
We employ three different voting methods, namely \textbf{Plurality Voting} \cite{plurality}, \textbf{Proportional Approval Voting (PAV)} \cite{joss-abcvoting} and \textbf{Ranked Choice Voting} 
\cite{Emerson2013} to determine the final summary. Due to the varying input requirements of different voting methods, changing both the prompting approach and the output generated by the LLM becomes imperative. 

Plurality voting and proportional voting are approval-based voting methods where voters can select multiple candidates they approve of without indicating a specific preference order.
In multi-winner plurality voting (also known as \textbf{block voting}), each voter casts multiple votes and the candidates are selected based on the number of votes polled. \textit{In the context of summarization, a textual unit is treated as a candidate, and the LLM acts as the voter}. We select the textual units in the decreasing order of the votes polled, till we obtain a summary of size $q$.
PAV evaluates the \textit{satisfaction} of each voter in the election outcome. A voter's satisfaction is measured based on -- amongst the number of candidates they voted for, how many are selected in the election. In the realm of summarization, PAV selects the textual units based on the amount of support each unit receives in $m$ summaries.
Since both plurality and proportional are approval-based voting algorithms, the units are 
either approved or disapproved by the underlying LLM, with no explicit ranking or preference order. In this case, we prompt the LLM to \textit{select the best $<q>$ sentences that summarize the input text} 
as shown in 
\circled{1}.

On the other hand, ranked choice voting entails assigning a score to each textual unit and subsequently selecting the highest-scoring units for inclusion in the summary. 
For ranked voting, we use the Borda Count, a positional voting algorithm \cite{Emerson2013}. In the Borda method, each candidate is assigned points corresponding to the number of candidates ranked below them: the lowest-ranked candidate receives 0 points, the next lowest gets 1 point, and so forth. The candidates with the highest aggregate points are declared as the winners.
In ranked voting, we prompt the LLM to \textit{output sentences in descending order of their suitability for the summary} as discussed in
\circled{2}.

It is important to note that the prompting technique and the output generated by LLM vary for different voting methods. In approval voting, the output from LLM is a list of $q$ textual units that LLM finds best suited to be included in the summary.
Whereas in ranked choice voting, the output from LLM is a list of the same length as input i.e. $s$ with all the units sorted in decreasing order of their preference towards the summary, 
and Borda Count~\cite{Emerson2013} is used to identify the top $q$ textual units. In the next section, 
we highlight how the voting-based summarization schemes outperform the {\it Vanilla} setup, which does not use voting. 
\section{Experimental Setup}

This section outlines the experimental setup, detailing the dataset utilized, the LLMs employed for summary generation, and the evaluation metrics used for assessment.

\subsection{Dataset}

\begin{table}[t]
\scriptsize
\centering
\begin{tabular}{l | l | r | r | r | r | r } 
\thickhline
\rowcolor{Gray} \textbf{No.} & \textbf{Post Category} & 
\textbf{\SBengaluru} & \textbf{\SChennai} & \textbf{\SDelhi} & \textbf{\SKolkata} & \textbf{\SMumbai} \\
\thickhline
PC1 & Rape/Sexual Assault & 33 & 27 & 17 & 23 & 18 \\
\hline

PC2 & Chain Snatching/Robbery & 49 & 16 & 103 & 32 & 22 \\ 
\hline

PC3 & Domestic Violence & 27 & 87 & 30 & 10 & 34 \\ \hline

PC4 & Physical Assault & 60 & 33 & 57 & 41 & 33\\ \hline

PC5 & Stalking & 146 & 166 & 165 & 100 & 153 \\ \hline

PC6 & Ogling/Staring & 147 & 100 & 202 & 133 & 209\\ 
\hline

PC7 & Taking Photos  &73 & 70 & 43 & 37 & 55 \\ 
\hline

PC8 & Mas*****tion in public & 45 & 55 & 27 & 30 & 42\\
\hline

PC9 & Touching/Groping & 152 & 209 & 206 & 159 & 230\\ \hline

PC10 & Showing Po**ography & 23 & 10 & 15 & 19 & 15 \\ 
\hline

PC11 & Commenting/Sexual Invites & 133 & 192 & 273 & 133 & 131 \\
\hline

PC12 & Online Harassment & 66 & 43 & 25 & 61 & 33 \\ \hline

PC13 & Human Trafficking & 3 & 4 & 2 & 2 & 1 \\ \hline

PC14 & Others & 50 & 23 & 36 & 41 & 16\\ \hline
\thickhline
\end{tabular}
\caption{The distribution of posts across each category for five datasets -- \City \SBengaluru, \SChennai, \SDelhi, \SKolkata, and \SMumbai. 
Posts tagged with various categories by their authors, with each category representing a different form of sexual harassment.}
\label{tab:post_count}
\end{table}

We gather the data from one of the major incident reporting platform in Asia. 
The dataset comprises incident posts from \textit{five} different cities, denoted as \City \SBengaluru, \SChennai, \SDelhi, \SKolkata and \SMumbai.
For \City \SDelhi and \SMumbai, we obtain the posts for 3 years, i.e., from Dec 2021 to Nov 2024. For \City \SBengaluru, \SChennai and \SKolkata, we consider the posts for 5 years, from Dec 2019 to Nov 2024.
The rationale behind varying the duration of post selection is to keep the total number of posts below 1000, facilitating more efficient and accurate annotation by the human summarizers. 
Our dataset of cities \SBengaluru, \SChennai, \SDelhi, \SKolkata and \SMumbai consists of 625, 867, 866, 545 and 728 posts respectively.
Sexual harassment can have various categories, such as physical assault, touching, stalking etc. 
Each post in the dataset is tagged with one or more categories by the author of the post.
The distribution of posts across these categories is shown in Table~\ref{tab:post_count}. While the platform does not collect personal information such as names or identities, it does gather age, gender, and details of the incidents. Table~\ref{tab:dataset_features} (in the Appendix) provides an overview of the attributes associated with each post. For our task, we focus solely on the main description provided in the posts.

For each of the five city-specific datasets, we generate gold-standard (reference) summaries created by three domain experts. These experts carefully selected textual units from the posts that are strong candidates for inclusion in the summary. As a result, each dataset has three expert-generated gold-standard summaries, each comprising 50 textual units (posts).
The expert annotators were provided with following guidelines for selecting the posts for the reference summary:

\begin{enumerate}
    \item Diversity: Prioritize diverse posts that represent various forms of assault.
    \item Descriptive: Give preference to posts with detailed descriptions over those containing only 2-3 words.
    \item Severity: Include posts that depict more serious cases or require urgent attention.
    \item Redundancy: Exclude posts that are repetitive or redundant.
\end{enumerate}

We utilize Fleiss Kappa to measure the inter-annotator agreement \cite{10.1037/h0031619}. The Fleiss Kappa scores for five datasets -- \City \SBengaluru, \SChennai, \SDelhi, \SKolkata and \SMumbai are 0.550, 0.376, 0.521, 0.543 and 0.446 respectively; showcasing moderate agreement between the annotators. Note that there may be multiple posts describing similar incidents, but different annotators might include different subsets of these posts in their summary.
Standard summary evaluation metrics like ROUGE (described in Section \ref{sec:metric}) consider these differences between individual reference summaries by averaging across multiple annotators.

\subsection{Large Language Models (LLMs)}
LLMs are characterized by their extensive parameter sizes and remarkable learning abilities \cite{zhao2023survey, chang2024survey}. 
In our work, we utilize two open LLMs: \texttt{llama-3.1-8B-instruct} from Meta \cite{grattafiori2024llama3herdmodels}, \texttt{open-mistral-nemo-2407} from Mistral AI \cite{mistral2024nemo}, and two proprietary LLMs:  \texttt{claude-3-haiku-20240307} from Claude \cite{TheC3} and \texttt{gpt-4o-mini-2024-07-18} from OpenAI \cite{gpt4omini}, to conduct experiments.

We focused on using models which are around 8B in parameter size. This choice was guided by practical considerations on the cost of computing infrastructure or API calls: open models of 8B size can conveniently run on GPUs with 40GB VRAM, unlike 70B models which typically require about 140GB VRAM. Similarly, 
API calls for smaller proprietary LLMs cost roughly one-tenth of their larger versions. For a non-profit platform, which may need to run these algorithms frequently, using larger and more expensive models would likely be impractical.

Across all experiments, we keep temperature, top probability and output tokens as 0, 1.0 and 8192 respectively. The current LLMs offer long context windows but we set the context window length to 8192 tokens to prevent hallucinations and ensure the LLMs adhere to the given instructions. 
Prior studies indicated that excessively long inputs can increase the likelihood of hallucination in LLM output \cite{zhang-etal-2023}. Empirically, we observed that providing long text often led to instruction neglect -- such as incomplete responses (e.g., only the first few sentences) or generic placeholders like `similarly we select other sentences'. In our experiments, limiting the context window to 8192 tokens resulted in minimal hallucination and more consistent compliance with the given instructions.

\subsection{Evaluation Metric}
\label{sec:metric}

To evaluate the quality of summaries generated by \ALGO, we employ evaluation metrics, namely ROUGE-1, ROUGE-2, and ROUGE-Lsum \cite{lin-2004-rouge}. These metrics quantify the degree of overlap between the \ALGO generated summary and the reference summary, thereby providing a quantitative measure of content preservation and relevance.

ROUGE-1 and ROUGE-2 are based on n-gram overlap, where ROUGE-1 measures the overlap of unigrams (individual words), and ROUGE-2 evaluates the overlap of bigrams (consecutive word pairs) between the generated and the reference summaries. These metrics capture the extent to which important words and short phrases from the reference summary are retained in the generated output, thereby reflecting content adequacy at a lexical level.

In contrast, ROUGE-L is based on the Longest Common Subsequence (LCS), which identifies the longest sequence of words that appear in both the candidate and reference summaries in the same order, though not necessarily contiguously. This property allows ROUGE-L to account for sentence-level structure and fluency.
Building on this, ROUGE-Lsum extends ROUGE-L by applying the LCS based matching at the sentence level rather than treating the summary as a single sequence. 
This formulation makes ROUGE-Lsum suitable for extractive summarization tasks, where the generated summary often consists of selected sentences from the source text. 
\section{Experimental Evaluation}\label{sec:results}

\begin{figure*}[t]
\centering
\includegraphics[clip, trim={0 0.2cm 0 0}, width=2.0\columnwidth]{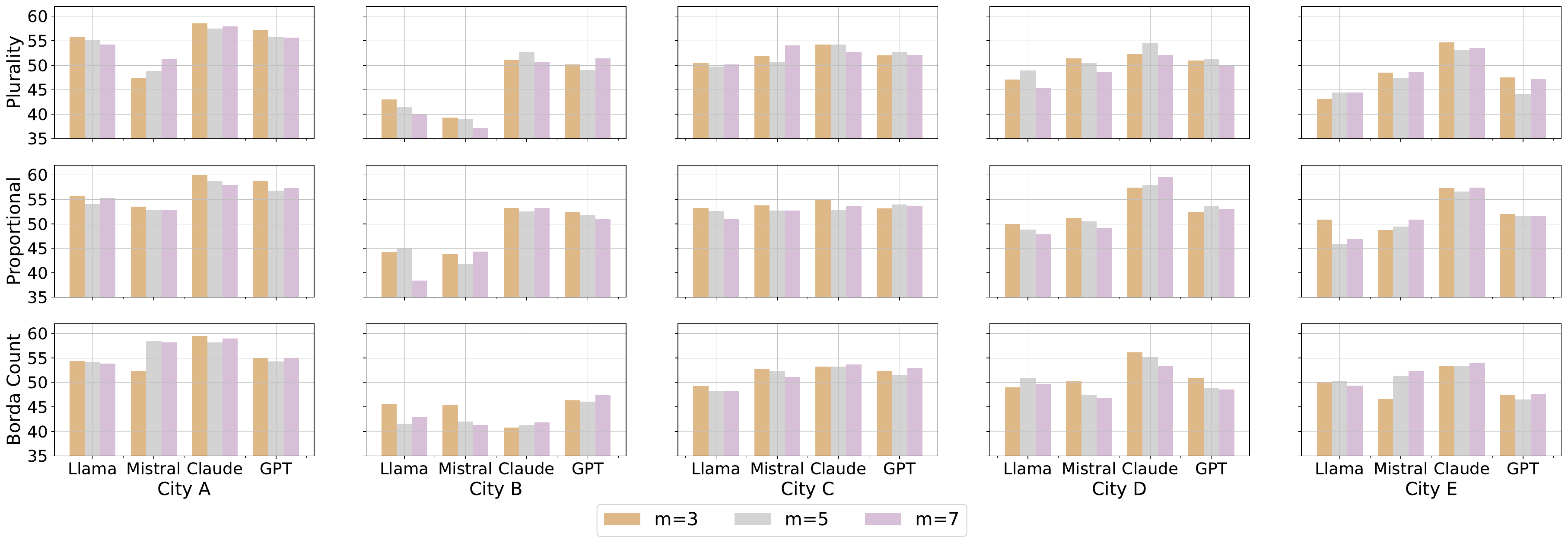}
\caption{Results for different values of $m$ for chunk size $s$ = 100. It can be observed that in most cases the results of $m$=3 are either better or comparable to the results of $m$=5 and $m$=7. Using $m$=3 proves to be more efficient for our experimental setup, offering advantages in both time and computational cost.}
\label{fig:ablation_s=100}
\end{figure*}

\begin{table}[t]
\small
\centering
\begin{tabular}{l | r | r | r | r | r} 
\thickhline
\rowcolor{Gray} \textbf{Parameters} & \textbf{\SBengaluru} & \textbf{\SChennai} & \textbf{\SDelhi} & \textbf{\SKolkata} & \textbf{\SMumbai} \\
\thickhline
\#textual units & 625 & 867 & 866 & 545 & 728 \\ \hline
\#words & 20544 & 12665 & 23501 & 22471 & 20807 \\  \hline
\#tokens & 30816 & 18997 & 35251 & 33706 & 31210 \\  
\thickhline
\end{tabular}
\caption{Number of textual units, words and tokens in five datasets.}
\label{tab:datasets}
\vspace{-1mm}
\end{table}

\begin{figure*}[t]
\centering
\includegraphics[clip, width=2\columnwidth]{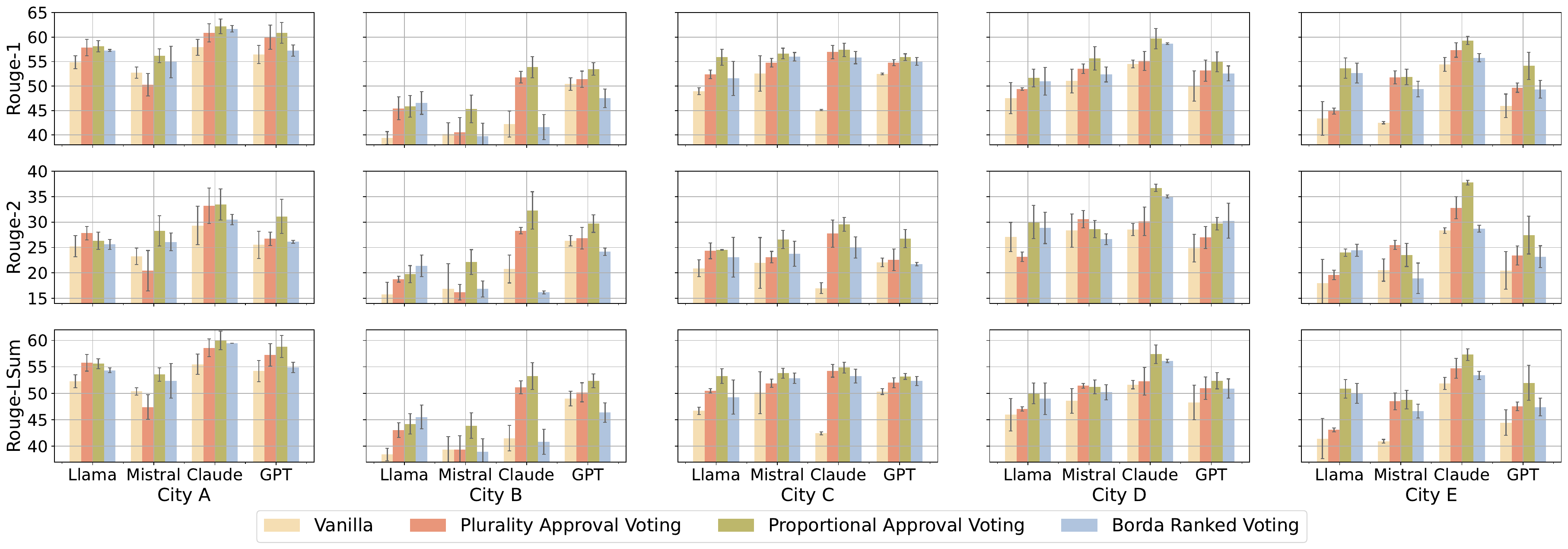}
\caption{Metric scores obtained through four different LLM setups. (i) Vanilla LLM without shuffling and voting method (ii) \ALGO with plurality approval voting (iii) \ALGO with proportional approval voting and (iv) \ALGO with borda count ranked voting. Results demonstrate that \texttt{claude-3-haiku} with proportional approval voting performs the best across all the cases. Here, Llama, Mistral, Claude and GPT refers to \texttt{llama-3.1-8B}, \texttt{open-mistral-nemo}, \texttt{claude-3-haiku} and \texttt{gpt-4o-mini} respectively. Error bars represent the standard deviation observed across 3 repeated runs of the same setup.} 
\label{fig:ablation}
\end{figure*}

In this section, we present the ablation studies, empirical comparison of \ALGO with competent baseline models and voting algorithms across datasets.
The total number of textual units ($|\TT|$), number of words, and number of tokens in each dataset are listed in Table \ref{tab:datasets}.
Value of $k$ (length of final summary) is set to 50 for all the experiments. 
To determine the optimal values of other hyperparameters such as $m$, $s$ and $q$, we conduct ablation studies, as discussed next.

\subsection{Ablation Study}
\label{sec:ablation}
\subsubsection{No. of Shuffles ($m$) and Chunk Size ($s$):} 
To identify the optimal number of shuffles $m$, we performed experiments using three values: 3, 5, and 7. Additionally, we tested two different chunk sizes $s$, set at 100 and 120. 
When the chunk size $s$ is 100, out of 60 cases, $m = 3$ gave superior results in 35 cases, $m = 5$ showed best performance in 12 cases, and $m = 7$ produced good results in 13 cases (refer Figure \ref{fig:ablation_s=100}).
For chunk size $s = 120$, out of 60 cases, $m = 3$, $m = 5$ and $m = 7$ showed good results in 24, 19 and 17 cases respectively (refer Figure \ref{fig:ablation_s=120} in Appendix). As the chunk size grows, more shuffles are needed as fewer shuffles may not sufficiently vary each post’s position within the input text. In our later experiments, we chose $m = 3$ and $s = 100$ since this combination consistently delivered the most reliable results.

\subsubsection{Which LLM Takes the Lead?}
We conducted experiments using four LLMs.
As illustrated in Figure \ref{fig:ablation}, \texttt{claude-3-haiku} outperformed the other three LLMs. 

\subsubsection{Summary Size from Chunk ($q$):} 
When $q \in [k,s)$, our proposed method is capable of handling the worst-case scenario where all textual units in the final summary may originate from a single level-0 chunk. As $q$ approaches $s$, a greater number of levels are necessary to reach the final summary. We hypothesize that the optimal value of $q$, which balances handling the worst-case scenario while minimizing the number of levels in multi-level summarization, is $k$. To validate this hypothesis, we experimented with five different values of $q$: 50, 60, 70, 80 and 90 on \Bengaluru dataset with Claude model. The results presented in Table 5 indicate that as $q$ increases, the number of levels needed also increases, while the ROUGE scores remain largely unaffected. Paired t-tests revealed that the differences were statistically insignificant, with p-values of 0.085, 0.72, 0.448, and 0.664 when comparing $q$ = 50 against $q$ = \{60, 70, 80, 90\}, respectively. Based on these findings, we selected $q$ = 50 as the optimal value, balancing performance and computational efficiency without compromising the results.


\begin{table}[t]
\small
\centering
\begin{tabular}{c | r | r | r | r} 
\thickhline
\rowcolor{Gray} \textbf{q} & \textbf{\#Levels} & \textbf{ROUGE-1} & \textbf{ROUGE-2} & \textbf{ROUGE-LSum} \\
\thickhline
50 &   4 &  60.364	&  31.988	&  57.875  \\
60 &   7 &  59.940	&  31.130	&  57.560  \\
70 &   9 &  60.871	&  30.122	&  58.276 \\
80 &  12 &  60.444	&  29.676	&  57.895 \\
90 &  24 &  61.132	&  28.946	&  58.304  \\
\thickhline
\end{tabular}
\caption{Ablation Study for different values of q for \Bengaluru using  \texttt{claude-3-haiku} model and proportional approval voting for $m$=3 and $s$=100.}
\label{tab:q_analysis}
\end{table}

\subsection{Baseline Comparison}

\begin{table*}[t]
\tiny
\centering
\begin{tabular}{l | c  c  c | c  c c | c  c c| c c c| c c c} 
\thickhline
\rowcolor{Gray}  & 
\multicolumn{3}{c|}{\textbf{\Bengaluru}} & 
\multicolumn{3}{c|}{\textbf{\Chennai}} &  
\multicolumn{3}{c|}{\textbf{\Delhi}} &  
\multicolumn{3}{c|}{\textbf{\Kolkata}} &  
\multicolumn{3}{c}{\textbf{\Mumbai}} \\
\cline{2-16}
\rowcolor{Gray}  \multirow{-2}{*}{\textbf{Models}} & 
\textbf{R1} & \textbf{R2} & \textbf{RLSum} & 
\textbf{R1} & \textbf{R2} & \textbf{RLSum} & 
\textbf{R1} & \textbf{R2} & \textbf{RLSum} & 
\textbf{R1} & \textbf{R2} & \textbf{RLSum} & 
\textbf{R1} & \textbf{R2} & \textbf{RLSum} \\
\thickhline
LexRank &  
57.644 & 28.084 & 55.904 & 
38.898 & 16.909	& 38.239 &
56.379 & 24.897	& 54.057 & 
46.221 & 21.709	& 44.612 & 
52.157 & 23.681 & 49.812 \\
SummBasic & 
52.498 & 21.848 & 50.700 &
26.714 &  9.536 & 26.283 &
53.172 & 21.600	& 51.136 &
43.215 & 19.841 & 41.392 &
50.677 & 21.357 & 48.482 \\
LSA & 
55.250 & 25.156 & 53.219 & 
52.529 & 25.947 & 51.462 & 
45.623 & 17.378 & 43.981 &
42.370 & 18.262 & 40.762 &
47.522 & 17.521 & 45.585 \\
BERT &
55.645 & 23.673 & 54.236 &
53.799 & 26.428 & 52.711 &
53.610 & 22.353 & 51.982 &
43.986 & 19.738 & 42.322 &
50.444 & 21.470 & 49.377 \\
XLNET & 
50.559 & 19.373 & 48.959 &
51.706 & 24.259 & 50.937 &
52.819 & 21.137 & 51.189 &
43.768 & 22.133 & 42.391 & 
51.650 & 22.151	& 50.142 \\
BERTSUM & 
56.570 & 25.323 & 54.226 &
51.137 & 24.551 & 49.386 &
53.927 & 22.647 & 50.993 &
53.767 & 25.846 & 49.376 &
47.735 & 20.606	& 45.295 \\ 
\ALGO  &  
\textbf{62.192} & \textbf{33.469}  & \textbf{59.996} &
\textbf{53.871}	& \textbf{32.273}  & \textbf{53.251} &
\textbf{57.401}	& \textbf{29.562}  & \textbf{54.854} &
\textbf{59.680}	& \textbf{36.721}  & \textbf{57.388} &
\textbf{59.324}	& \textbf{37.784}  & \textbf{57.312} \\
\thickhline
\end{tabular}
\caption{Table showing metric scores from different models for various datasets. Here, R1 = ROUGE-1 Score, R2 = ROUGE-2 Score, RLSum = ROUGE-LSum Score. The best result for each dataset is shown in \textbf{bold} and clearly \ALGO outperforms all the other methods across all the evaluation measures.}
\label{tab:results}
\end{table*}

We compare \ALGO with the 
pre-neural models such as 
LexRank \cite{Erkan:2004}, SummBasic \cite{Nenkova05theimpact} and LSA \cite{Gong:2001};
transformer based models such as BERT \cite{miller2019leveraging} and XLNET \cite{10.5555/3454287.3454804}; state-of-the-art fine-tuned model BERTSUM \cite{liu-lapata-2019-text}. 
Our proposed method, \ALGO, achieves optimal performance when \texttt{claude-3-haiku} is used with proportional approval voting. Therefore, we present the results of this combination as the outcome of \ALGO. 
As shown in Table \ref{tab:results}, it is observed that \ALGO surpasses state-of-the-art summarization models across all metrics. 


\subsection{Does \ALGO Perform Better than Vanilla LLM?}

\label{subsec:vanilla_vs_lamsum}
Our proposed framework, \ALGO, ensures robust summary generation by shuffling and employing a voting algorithm to select the best textual units for the summary. 
It is crucial to compare \ALGO with a multi-level LLM that does not use shuffling and voting, which we call \textit{Vanilla LLM}. Algorithm \ref{algo:vanilla} 
outlines the steps used by \textit{vanilla LLM} to find the chunk summary. 
Figure \ref{fig:ablation} demonstrates that the \textit{vanilla} multi-level LLM has lower ROUGE scores for each LLM compared to the proposed framework \ALGO, indicating that shuffling and voting enhance the performance.
Earlier work \cite{zhang-etal-2023} reported that the ChatGPT model achieves lower ROUGE scores on CNN/DM and XSum datasets.
Our results demonstrate that our proposed framework performs significantly better than other fine-tuned language models such as BERTSUM for large user-generated text. 

\subsection{Which Voting Algorithm Performs the Best?}
We experimented with three voting algorithms, two approval-based (plurality and proportional) and one rank-based (borda-count). 
Experimental results indicate that LLMs with proportional approval voting perform the best compared to the other voting algorithms (Figure \ref{fig:ablation}). 
Differences in performance for different voting algorithms were found to be statistically significant (p $<$ 0.05 in paired t-test) for all the setups, except Mistral with plurality based voting and GPT with borda count (refer Table \ref{tab:pvalue} in Appendix).

We hypothesized that rank-based voting would yield better results, as it makes more informed decisions about the potential sentences to be included in the summary. Contrary to our expectations, rank-based algorithms did not surpass the proportional approval voting. This can be attributed to multiple factors: 
\begin{enumerate}
    \item LLMs may hallucinate and output sentences in the same or in the reverse order as they were in the input.
    \item Occasionally, LLMs do not output all the sentences from the input, resulting in the padding of left-out sentences towards the end of the list, which disturbs the ranking and potentially affects the result.
\end{enumerate}

Proportional Approval Voting (PAV) selects posts in proportion to the support posts receive from different shuffles. Often, several posts express similar ideas -- such as repeating a fact, argument, or opinion. There is a risk of redundancy if we select the posts with the highest number of approvals -- the summary may emphasize the same point multiple times across different shuffles, while overlooking less frequent but important posts.
PAV addresses this by applying diminishing returns. A post gains credit not just based on how many shuffles approved it, but also based on how much utility it provides to each shuffle. For each shuffle, the satisfaction decreases with each additional approved post selected -- the first approved post contributes more than the second, the second more than the third, and so on.
This approach helps avoid over representation of dominant viewpoints and encourages inclusion of varied content. PAV behaves like an editor that takes into account all shuffles, detects overlaps, and constructs a summary that balances popular content with distinct, less common insight -- capturing the collective judgment more effectively.


\begin{algorithm}[t]
\small
\caption{Algorithm for summarization of a chunk in Vanilla LLM}\label{algo:vanilla}
\begin{algorithmic}
\Function{ChunkResult}{$\TT, si, ei, q, m$}
\State $R = \textproc{LLM}(\TT,si,ei,q)$ \Comment{$q$ textual units from $[si,ei]$}
\State $C = \textproc{Check}(R,\TT,si,ei)$ \Comment{output calibration}
\State return $C$ 
\EndFunction
\end{algorithmic}
\end{algorithm}



\subsection{Analyzing the \ALGO Output}

\begin{figure}[t]
\centering
\includegraphics[clip, width=0.9\columnwidth]{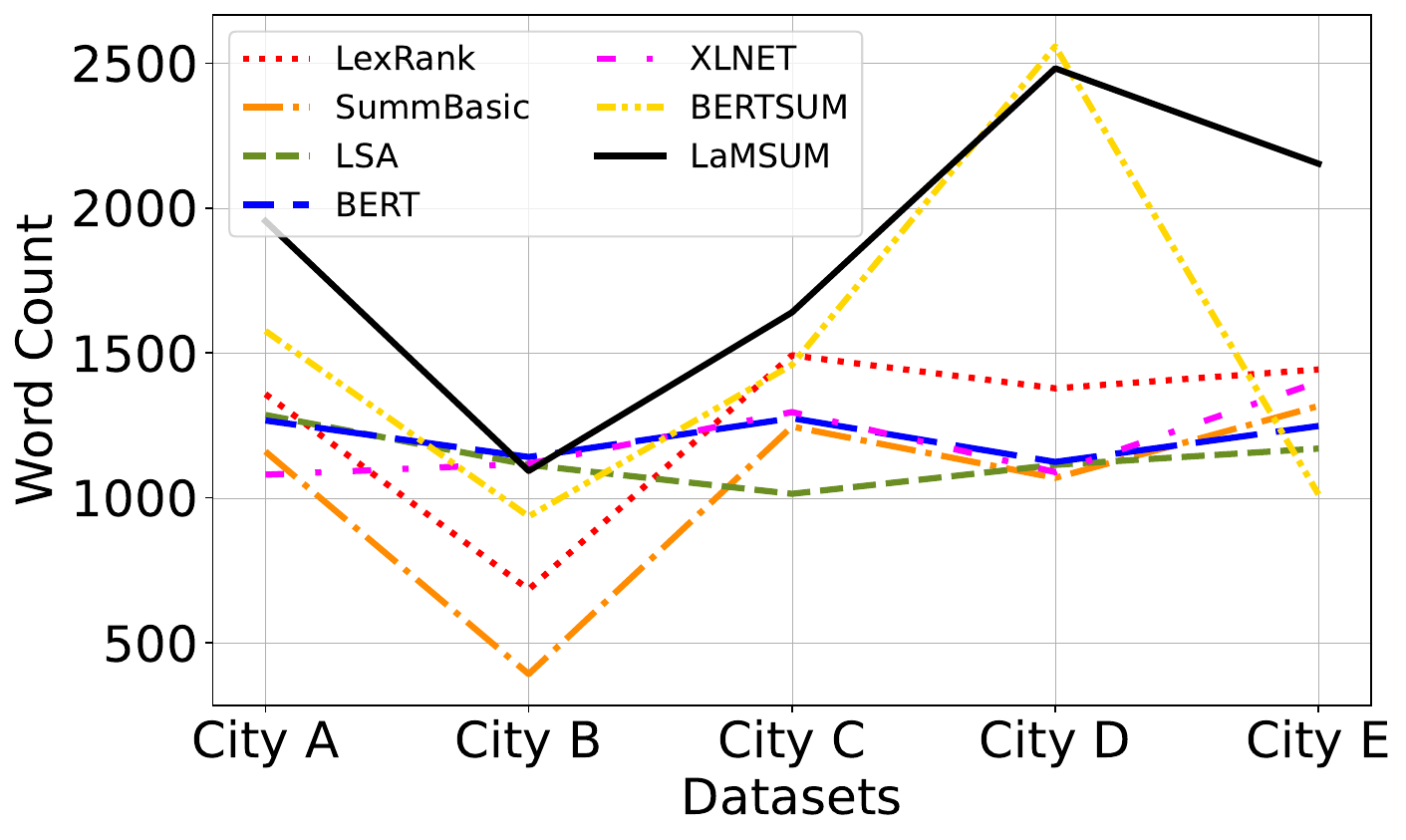}
\caption{Posts chosen by \ALGO tend to be detailed and descriptive, offering a deeper level of information. Number of words in \ALGO selected posts is often highest across various datasets, ensuring extensive and comprehensive summarization.} 
\label{fig:words_as}
\end{figure}

\begin{table}[t]
\scriptsize
\centering
\begin{tabular}{l | r  r  r  r  r} 
\thickhline
\rowcolor{Gray} Models & 
{\textbf{\Bengaluru}} & 
{\textbf{\Chennai}} &  
{\textbf{\Delhi}} &  
{\textbf{\Kolkata}} &  
{\textbf{\Mumbai}} \\
\thickhline
LexRank &  7.486 & 5.819 & 7.637 & 7.548 & 7.481\\
SummBasic &  8.198 & 5.734 & 8.050 & 8.020 & 8.196\\
LSA & 8.251 & 7.061 & 8.481 & 8.194 & 8.387\\
BERT & 8.068 & 7.762 & 8.437 & 8.186 &  8.191\\
XLNET & 8.144 & 7.689 & 8.619 & 8.205 & 8.400 \\
BERTSUM & 8.331 & 7.539 & 8.690 & 7.960 & 8.112 \\ 
\ALGO  & \textbf{8.563} & \textbf{7.822} & \underline{8.622} & \textbf{8.606} & \textbf{8.480} \\ 
\thickhline
\end{tabular}
\caption{Entropy values representing the diversity in the summaries produced by various algorithms. \textbf{Bold} values highlight the best and the \underline{underline} denotes the second-best performance. \ALGO achieves the highest diversity score across four datasets. 
}
\label{tab:entropy}
\end{table}

We analyse the difference between the posts selected by the \ALGO and those chosen by the other algorithms. \ALGO selects the posts that are more descriptive and rich in detail, in contrast to posts that lack sufficient information. 
As shown in Figure \ref{fig:words_as}, it is evident that across nearly all datasets, the posts selected by \ALGO exhibit a higher word count. This indicates that the proposed algorithm is capable of capturing more detailed information compared to the other algorithms (refer Table \ref{tab:sample_posts} in Appendix for examples). Furthermore, posts selected by \ALGO exhibit greater diversity, encompassing a broader range of harassment categories compared to the other baselines. We use entropy as a measure of diversity where a higher value is indicative of more randomness \cite{entropy}. Table \ref{tab:entropy} 
demonstrates that \ALGO generates diverse summaries as compared to other baselines.

\section{Concluding Discussion}
Incident reporting platforms receive numerous posts related to sexual harassment; a summarization algorithm enables end-users to quickly review the most significant ones.
This work marks an early attempt to achieve extractive summarization of large user-generated text that exceeds a single context window using zero-shot learning. 
The proposed multi-level framework \ALGO leverages approval based and ranked-based voting algorithms to generate robust summaries. 
Experiments conducted on crowd-sourced dataset demonstrated the efficacy of \ALGO, as it outperformed the results achieved by state-of-the-art models. 
While our primary focus was on applying \ALGO to an incident reporting platform, the proposed framework is generalizable and can be readily used with other social media datasets. We demonstrated this by evaluating \ALGO on three publicly available datasets -- US-Election, Claritin, and Me-Too \cite{dash2018fairsumm}. Additional details on the datasets and results can be found in Appendix Section \ref{sec:generalizability}. 
The overall contribution of this study is a socially grounded framework for extractive summarization of user-reported harassment incidents and the use of LLMs for socially responsible applications.

Note that there can be a concern regarding the potential data leakage, as the experiments involve newer LLMs that may have been exposed to the experimental datasets during their pre-training phase. 
We demonstrate that the \textit{vanilla LLM}, despite being an LLM-based framework, exhibits inferior performance, whereas our proposed framework generates more robust summaries and delivers improved results.
This highlights the efficacy of our model, even when it is exposed to data leakage. 

\vspace{1mm}
\noindent \textbf{Limitation:}
Our proposed framework, \ALGO, very well handles text of any length, conditioned on the fact that the final summary fits within a single context window.
Some modifications to \ALGO may be necessary when the output summary exceeds the size of a single context window.

\vspace{1mm}
\noindent \textbf{Ethical Considerations:}
Our research focuses on using LLMs to produce extractive summaries for incident reporting platforms. 
LLMs often exhibit bias towards their training data~\cite{chhikara2024few}, which can influence their preference for certain textual elements during the summarization process.
Their ``black box'' nature, with an opaque decision-making process, makes it challenging to discern how or why specific textual units are chosen for summarization. 
The posts selected by \ALGO may not accurately represent real-life scenarios and cannot serve as a reliable proxy for actual situations. Additionally, \ALGO may overlook less frequent posts with limited informational content. As a result, exclusive reliance on our framework could lead to the oversight of specific issues by the authorities.
While \ALGO can produce high-quality summaries, its use must be approached with careful consideration of potential ethical implications.

\section*{Acknowledgments}
We would like to express our sincere gratitude to the anonymous reviewers for their valuable feedback which helped in considerably improving the quality of the paper. 

\bibliography{ref}
\clearpage
\section*{Checklist}

\begin{enumerate}

\item For most authors...
\begin{enumerate}
    \item  Would answering this research question advance science without violating social contracts, such as violating privacy norms, perpetuating unfair profiling, exacerbating the socio-economic divide, or implying disrespect to societies or cultures?
    \answerYes{Yes}
  \item Do your main claims in the abstract and introduction accurately reflect the paper's contributions and scope?
    \answerYes{Yes}
   \item Do you clarify how the proposed methodological approach is appropriate for the claims made? 
    \answerYes{Yes}
   \item Do you clarify what are possible artifacts in the data used, given population-specific distributions?
    \answerYes{Yes}
  \item Did you describe the limitations of your work?
    \answerYes{Yes}
  \item Did you discuss any potential negative societal impacts of your work?
    \answerYes{Yes}
  \item Did you discuss any potential misuse of your work?
    \answerYes{Yes}
    \item Did you describe steps taken to prevent or mitigate potential negative outcomes of the research, such as data and model documentation, data anonymization, responsible release, access control, and the reproducibility of findings?
    \answerYes{Yes}
  \item Have you read the ethics review guidelines and ensured that your paper conforms to them?
    \answerYes{Yes}
\end{enumerate}

\item Additionally, if your study involves hypotheses testing...
\begin{enumerate}
  \item Did you clearly state the assumptions underlying all theoretical results?
    \answerNA{NA}
  \item Have you provided justifications for all theoretical results?
    \answerNA{NA}
  \item Did you discuss competing hypotheses or theories that might challenge or complement your theoretical results?
    \answerNA{NA}
  \item Have you considered alternative mechanisms or explanations that might account for the same outcomes observed in your study?
    \answerNA{NA}
  \item Did you address potential biases or limitations in your theoretical framework?
    \answerNA{NA}
  \item Have you related your theoretical results to the existing literature in social science?
    \answerNA{NA}
  \item Did you discuss the implications of your theoretical results for policy, practice, or further research in the social science domain?
    \answerNA{NA}
\end{enumerate}

\item Additionally, if you are including theoretical proofs...
\begin{enumerate}
  \item Did you state the full set of assumptions of all theoretical results?
    \answerNA{NA}
	\item Did you include complete proofs of all theoretical results?
    \answerNA{NA}
\end{enumerate}

\item Additionally, if you ran machine learning experiments...
\begin{enumerate}
  \item Did you include the code, data, and instructions needed to reproduce the main experimental results (either in the supplemental material or as a URL)?
    \answerYes{Yes}
  \item Did you specify all the training details (e.g., data splits, hyperparameters, how they were chosen)?
    \answerYes{Yes}
     \item Did you report error bars (e.g., with respect to the random seed after running experiments multiple times)?
    \answerYes{Yes}
	\item Did you include the total amount of compute and the type of resources used (e.g., type of GPUs, internal cluster, or cloud provider)?
    \answerYes{Yes}
     \item Do you justify how the proposed evaluation is sufficient and appropriate to the claims made? 
    \answerYes{Yes}
     \item Do you discuss what is ``the cost'' of misclassification and fault (in)tolerance?
    \answerNA{NA}
  
\end{enumerate}

\item Additionally, if you are using existing assets (e.g., code, data, models) or curating/releasing new assets, \textbf{without compromising anonymity}...
\begin{enumerate}
  \item If your work uses existing assets, did you cite the creators?
    \answerYes{Yes}
  \item Did you mention the license of the assets?
    \answerYes{Yes}
  \item Did you include any new assets in the supplemental material or as a URL?
    \answerYes{Yes}
  \item Did you discuss whether and how consent was obtained from people whose data you're using/curating?
    \answerNA{NA}
  \item Did you discuss whether the data you are using/curating contains personally identifiable information or offensive content?
    \answerYes{Yes}
\item If you are curating or releasing new datasets, did you discuss how you intend to make your datasets FAIR (see \citet{fair})?
\answerYes{Yes}
\item If you are curating or releasing new datasets, did you create a Datasheet for the Dataset (see \citet{gebru2021datasheets})? 
\answerNA{NA}
\end{enumerate}

\item Additionally, if you used crowdsourcing or conducted research with human subjects, \textbf{without compromising anonymity}...
\begin{enumerate}
  \item Did you include the full text of instructions given to participants and screenshots?
    \answerYes{Yes}
  \item Did you describe any potential participant risks, with mentions of Institutional Review Board (IRB) approvals?
    \answerNA{NA}
  \item Did you include the estimated hourly wage paid to participants and the total amount spent on participant compensation?
    \answerNA{NA}
   \item Did you discuss how data is stored, shared, and deidentified?
   \answerYes{Yes}
\end{enumerate}

\end{enumerate}
\clearpage
\appendix
\section{Appendix}
\label{sec:appendix}

\subsection{Algorithm}

Algorithm \ref{algo:calibration} shows the method for output calibration by utilizing two modules -- i). minimum edit distance and ii). maximum count of keywords.

\begin{algorithm}[H]
\caption{Algorithm for output calibration}
\label{algo:calibration}
\begin{algorithmic}
\Function{Check}{$R,\TT,si,ei$}
\State $Y = \{\}$ \Comment{store the result}
\For{$x$ in $R$} \Comment{for each sentence $x$ in LLM result $R$}
\State $min\_dist = \infty$ \Comment{keep track of min distance}
\State $min\_idx = -1$ \Comment{post with min edit distance}
\State $max\_count = 0$ \Comment{matching keywords}
\State $max\_idx = -1$ \Comment{post with max keywords}
\State $K = \textproc{Keywords}(x)$ \Comment{obtain keywords in $x$}
\For{$i \gets si$ to $ei$} \Comment{for each unit in \TT}
\State $d = \textproc{EditDist}(x, \TT[i])$ \Comment{obtain edit distance}
\If{$d < min\_dist$} \Comment{lesser edit distance}
\State $min\_dist = d$ \Comment{update $min\_dist$}
\State $min\_idx = i$ \Comment{update $min\_idx$}
\EndIf
\State $c = \textproc{Count}(K, \TT[i])$ \Comment{\# keywords in $\TT[i]$}
\If{$c > max\_count$} \Comment{lesser edit distance}
\State $max\_count = c$ \Comment{update $max\_count$}
\State $max\_idx = i$ \Comment{update $max\_idx$}
\EndIf
\EndFor
\If{$min\_dist < \epsilon$} \Comment{edit distance is low}
\State $Y.add(\TT[min\_idx])$ \Comment{add to result $Y$}
\Else
\State $Y.add(\TT[max\_idx])$ \Comment{add to result $Y$}
\EndIf
\EndFor
\State return $Y$
\EndFunction
\end{algorithmic}
\end{algorithm}

\subsection{Zero Shot Prompting}
\begin{table}[H]
\small
\centering
\begin{tabular}{p{8.10cm}} 
\thickhline
\rowcolor{Gray} \textbf{Prompts} \\
\thickhline
\circled{1} 
Select the most suitable units that summarize the input text. \\ 
\hline
Prompt: Input consists of $<$chunk\_size$>$ sentences. Each sentence is present in a new line. Each sentence contains a sentence number followed by text. You are an assistant that selects best $<$summary\_length$>$ sentences (subset) which summarizes the input. Think step by step and follow the instructions. $<$sentences$>$ \\
\thickhline
\circled{2} Generate a ranked list in descending order of preference. \\
\hline
Prompt: Input consist of $<$chunk\_size$>$ sentences. Each sentence is present in a new line. Each sentence contains a sentence number followed by text. You are an assistant that outputs the sentences in the decreasing order of their relevance to be included in the summary. Remember that output should contain all the sentences in the decreasing order of their relevance. $<$sentences$>$ \\
\thickhline
\end{tabular}
\caption{Prompts utilised for \circled{1} Approval and \circled{2} Ranked-based voting algorithm.}
\label{tab:prompts}
\end{table}



\subsection{What Fails to Deliver Extractive Summary?}
To ensure extractive summarization, we tested an additional approach -- each sentence is tagged with a sentence number, LLM is prompted to \textit{select the best $q$ sentences and output only the sentence numbers of the best $q$ sentences}. 
Thereafter, the sentences corresponding to the sentence numbers can be retrieved. 
For instance, if $s$ is 100 and $q$ is 50, 
the task is to output the sentence numbers of the best 50 sentences from a pool of 100 sentences.
In such cases, LLMs hallucinate and provide an output consisting of either all the odd number sentences or all the even number sentences. 

\noindent {\bf Takeaway:} For extractive summarization, relying solely on indexes may result in hallucination, underscoring the importance of emitting the input content and not the numbers. 

\subsection{Post Features}
\begin{table}[H]
\small
\centering
\begin{tabular}{l | l} 
\thickhline
\rowcolor{Gray} \textbf{Feature} & \textbf{Details} \\
\thickhline
id	& unique id for each post\\
lang\_id & language	id\\
building & building where incident took place\\
landmark & landmark near the place of incident\\
area & area where incident occurred\\
city & city where incident happened \\
state & name of the state\\
country & country name\\
latitude & coordinates information \\
longitude & coordinates information \\
created\_on	& date when the post is made\\
\underline{description}	& details about the incident\\
additional\_detail & more information about the incident\\
age	& age of the person\\
gender\_id & gender id\\
gender	& gender of the person \\
incident\_date & date when the incident took place\\
is\_date\_estimate & binary value -- yes or no\\
time\_from & start time of the incident	\\
time\_to & end time of the incident \\
is\_time\_estimate & binary value -- yes or no \\
categories	& harassment category\\
\thickhline
\end{tabular}
\caption{Features or attributes associated with a post. For our task, we utilise only the \underline{description} feature.}
\label{tab:dataset_features}
\end{table}

\subsection{Generalizability to Other Datasets}
\label{sec:generalizability}
We run our experiments on {\it three} publicly available datasets \cite{dash2018fairsumm}. \textit{Claritin} dataset contains 4,037 tweets about the benefits and the side-effects of the anti-allergic drug Claritin. \textit{US-Election} dataset contains 2,120 tweets from 2016 US Presidential Election where people support and attack different political parties. \textit{Me-Too} dataset includes 488 tweets from the October 2018 MeToo movement, where individuals recount the harassment cases they experienced. Results for these datasets are shown in Table \ref{tab:more_dataset_results}. Results demonstrate that \ALGO delivers the best results across all baseline algorithms.

\begin{table*}[t]
\small
\centering
\begin{tabular}{ l | c  c  c | c  c  c | c  c  c } 
\thickhline
\rowcolor{Gray} & 
\multicolumn{3}{c|}{\textbf{Claritin}} & 
\multicolumn{3}{c|}{\textbf{US-Election}} &  
\multicolumn{3}{c}{\textbf{MeToo}} \\
\cline{2-10}
\rowcolor{Gray} \multirow{-2}{*}{\textbf{Models}}  & \textbf{R1} & \textbf{R2} & \textbf{RLSum} & \textbf{R1} & \textbf{R2} & \textbf{RLSum} & \textbf{R1} & \textbf{R2} & \textbf{RLSum} \\
\thickhline
LexRank     & 45.04 & 19.71 & 44.74 & 42.63 & 10.78 & 41.64 & 42.70 & 11.32 & 40.91 \\
SummBasic   & 58.25 & 19.29 & 56.76 & 55.36 & 12.43 & 53.94 & 57.23 & 18.53 & 54.07 \\
LSA      & 61.61 & 23.58 & 60.74 & 55.86 & 15.07 & 54.81 & 40.63 & 11.24 & 38.86 \\
BERT        & 57.30 & 22.37 & 56.21 & 55.89 & 15.44 & 55.00 & 45.72 & 10.76 & 43.50 \\
XLNET       & 55.52 & 21.37 & 54.75 & 56.48 & 15.72 & 55.41 & 36.58 & 08.50 & 34.48 \\
BERTSUM     & 57.87 & 22.75 & 55.96 & 59.00 & 17.51 & 57.41 & 57.11 & 17.08 & 54.84 \\ 
\ALGO & \textbf{64.20} & \textbf{26.71} & \textbf{62.66} & \textbf{60.11} & \textbf{18.26} & \textbf{58.99} & \textbf{58.14} & \textbf{21.99} & \textbf{55.46} \\
\thickhline
\end{tabular}
\caption{Metric scores from different models for various datasets. Here, R1 = ROUGE-1 Score, R2 = ROUGE-2 Score, RLSum = ROUGE-LSum Score. The best value per evaluation measure is shown in \textbf{bold} and we can observe that \ALGO outperforms the baseline models.}
\label{tab:more_dataset_results}
\end{table*}

\begin{figure*}[t]
\centering
\includegraphics[clip, trim={0 0.2cm 0 0}, width=2.0\columnwidth]{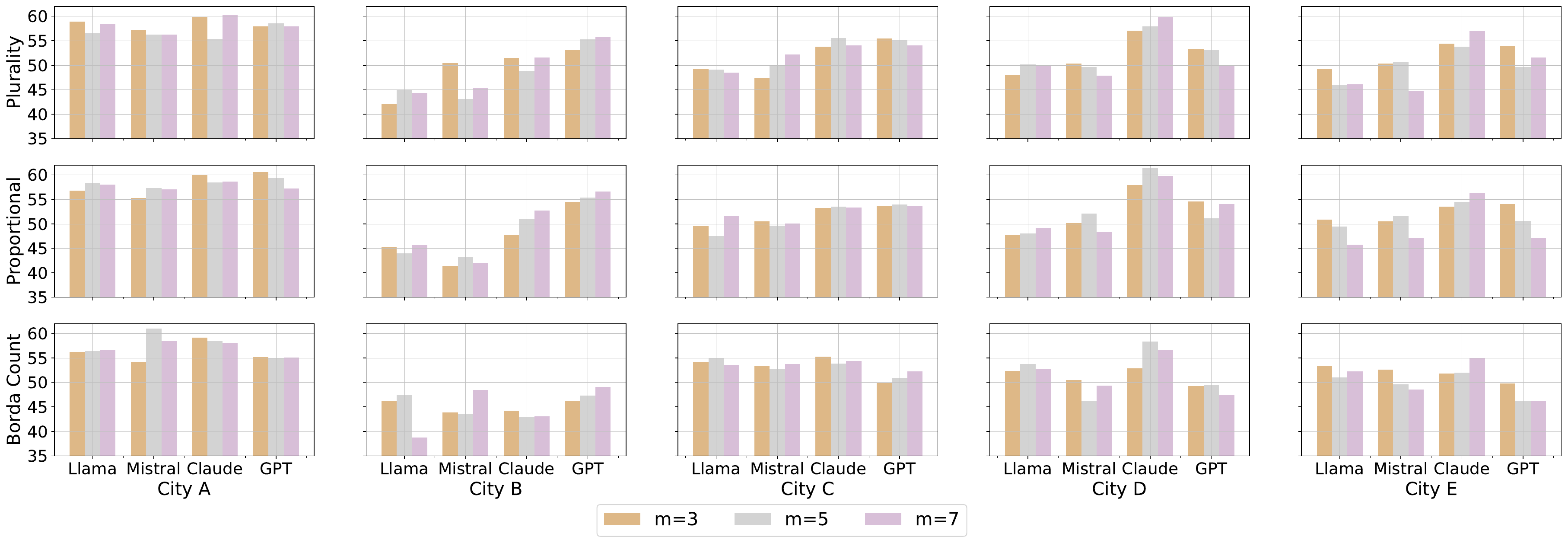}
\caption{Results for different values of $m$ with a chunk size of $s = 120$ show that out of 60 cases, $m = 3$, $m = 5$, and $m = 7$ achieved good performance in 24, 19, and 17 cases respectively. As the chunk size increases, performing only three shuffles may be insufficient to guarantee that each post occupies varied positions within the input text.}
\label{fig:ablation_s=120}
\end{figure*}

\begin{table*}[t]
\small
\centering
\begin{tabular}{l | r | r | r} 
\thickhline
\rowcolor{Gray} \textbf{Model} & \textbf{Plurality} &  \textbf{Proportional}  & \textbf{Borda Count}\\
\thickhline
Llama & 0.00061722 & 0.00000123 & 0.00003112  \\
Mistral & \textcolor{red}{0.08018935} & 0.00000118 & 0.01398864 \\
Claude & 0.00011471 & 0.00000026 & 0.00855709 \\
GPT & 0.00000225 & 0.00000002 & \textcolor{red}{0.07184720} \\
\thickhline
\end{tabular}
\caption{The p-values from the paired t-test indicate the statistical significance of the results when comparing the vanilla setup to \ALGO using various voting algorithms. All comparisons showed statistically significant results with p $<$ 0.05, except for for Mistral with plurality voting and for GPT with borda count.}
\label{tab:pvalue}
\end{table*}

\begin{table*}[t]
\tiny
\centering
\begin{tabular}{l | c  c c | c  c c | c  c c | c c c | c c c} 
\thickhline
\rowcolor{Gray}  & 
\multicolumn{3}{c|}{\textbf{\Bengaluru}} & 
\multicolumn{3}{c|}{\textbf{\Chennai}} &  
\multicolumn{3}{c|}{\textbf{\Delhi}} &  
\multicolumn{3}{c|}{\textbf{\Kolkata}} &  
\multicolumn{3}{c}{\textbf{\Mumbai}} \\
\cline{2-16}
\rowcolor{Gray} \multirow{-2}{*}{\textbf{Models}} & 
\textbf{R1} & \textbf{R2} & \textbf{RLSum} & 
\textbf{R1} & \textbf{R2} & \textbf{RLSum} & 
\textbf{R1} & \textbf{R2} & \textbf{RLSum} & 
\textbf{R1} & \textbf{R2} & \textbf{RLSum} & 
\textbf{R1} & \textbf{R2} & \textbf{RLSum} \\
\thickhline
\rowcolor{Gray} \multicolumn{16}{c}{Vanilla LLM} \\ 
\thickhline
Llama & 
54.860	& 25.240  & 52.304  &
39.391	& 15.795  & 38.430  &
48.935	& 20.895  & 46.677  &
47.528	& 27.047  & 45.946  &
43.416	& 17.988  & 41.419 \\
Mistral & 
52.750	& 23.234  & 50.362  &
40.181	& 16.885  & 39.341  &
52.547	& 21.972  & 50.113  &
51.037	& 28.340  & 48.561  &
42.535	& 20.544  & 40.919 \\
Claude  &
57.913	& 29.320  & 55.496  &
42.209	& 20.780  & 41.502  &
45.103	& 17.000  & 42.434  &
54.525	& 28.542  & 51.617  &
54.413	& 28.326  & 51.855 \\
GPT & 
56.458	& 25.519  & 54.214  &
50.430	& 26.350  & 49.008  &
52.488	& 22.053  & 50.336  &
50.011	& 24.889  & 48.246  &
45.980	& 20.473  & 44.439 \\
\thickhline
\rowcolor{Gray} \multicolumn{16}{c}{\ALGO + Plurality Voting} \\ 
\thickhline
Llama & 
57.876	& 27.809  & 55.756  &
45.459	& 18.787  & 43.034  &
52.376	& 24.352  & 50.468  &
49.407	& 23.182  & 47.072  &
44.879	& 19.623  & 43.080 \\
Mistral & 
50.258	& 20.426  & 47.387  &
40.544	& 16.201  & 39.317  &
54.749	& 23.093  & 51.887  &
53.531	& 30.556  & 51.430  &
51.777	& 25.503  & 48.494 \\
Claude & 
60.896	& 33.200  & 58.597  &
51.807	& 28.308  & 51.109  &
56.950	& 27.725  & 54.239  &
55.130	& 30.145  & 52.282  &
57.361	& 32.805  & 54.699 \\
GPT  & 
59.992	& 26.724  & 57.271  &
51.400	& 26.819  & 50.183  &
54.771	& 22.595  & 52.002  &
53.160	& 26.968  & 50.944  &
49.675	& 23.408  & 47.548\\
\thickhline
\rowcolor{Gray} \multicolumn{16}{c}{\ALGO + Proportional Voting} \\ 
\thickhline
Llama & 
58.136	& 26.325  & 55.590  &
45.844	& 19.763  & 44.208  &
55.884	& 24.522  & 53.271  &
51.655	& 30.022  & 49.956  &
53.669	& 23.998  & 50.840 \\
Mistral & 
56.202	& 28.285  & 53.548  &
45.298	& 22.138  & 43.887  &
56.648	& 26.570  & 53.803  &
55.639	& 28.618  & 51.206  &
51.889	& 23.516  & 48.772 \\
Claude & 
\textbf{62.192} & \textbf{33.469}  & \textbf{59.996} &
\textbf{53.871}	& \textbf{32.273}  & \textbf{53.251} &
\textbf{57.401}	& \textbf{29.562}  & \textbf{54.854} &
\textbf{59.680}	& \textbf{36.721}  & \textbf{57.388} &
\textbf{59.324}	& \textbf{37.784}  & \textbf{57.312} \\
GPT &
60.872	& 31.103  & 58.844  &
53.489	& 29.688  & 52.356  &
55.901	& 26.762  & 53.179  &
54.965	& 29.699  & 52.366  &
54.121	& 27.445  & 51.981 \\
\thickhline
\rowcolor{Gray} \multicolumn{16}{c}{\ALGO + Borda Count}\\ 
\thickhline
Llama & 
57.262	& 25.622  & 54.349 &
46.561	& 21.386  & 45.518 &
51.584	& 23.077  & 49.278 &
51.009	& 28.853  & 48.961 &
52.670	& 24.440  & 49.949 \\
Mistral &  
54.928	& 26.103  & 52.353  &
39.762	& 16.865  & 38.956  &
55.995	& 23.740  & 52.834  &
52.379	& 26.646  & 50.213  &
49.389	& 18.947  & 46.623 \\
Claude &
61.693	& 30.466  & 59.470   & 
41.650	& 16.165  & 40.822  &
55.797	& 25.026  & 53.268  &
58.646	& 35.064  & 56.132  &
55.788	& 28.688  & 53.387 \\
GPT & 
57.244	& 26.107  & 54.881  &
47.500	& 24.153  & 46.348  &
55.031	& 21.72	  & 52.331  &
52.608	& 30.263  & 50.900  &
49.323	& 23.184  & 47.390 \\
\thickhline
\end{tabular}
\caption{Table showing metric scores from different LLM models for various datasets. The best value per dataset is shown in \textbf{bold} and clearly \texttt{claude-3-haiku} with proportional approval voting outperforms all the other methods across all the evaluation measures. 
In this table, Llama, Mistral, Claude and GPT refers to \texttt{llama-3.1-8B}, \texttt{open-mistral-nemo}, \texttt{claude-3-haiku} and \texttt{gpt-4o-mini} respectively. Graphical representation of this table is shown in Figure \ref{fig:ablation}.}
\label{tab:ablation}
\end{table*}

\begin{table*}[t]
\small
\centering
\begin{tabular}{p{17cm}} 
\thickhline
\rowcolor{Gray} \textbf{Posts} \\
\thickhline
One person whom my family rejected for marriage is posting my nude pictures in social media platforms, hacked my email id and forwarding the nude videos and pictures to all the contacts through different different email ids. Fake Facebook id using my pics and also fake Instagram accounts using my pics, posting bad things about me and my mother. Every day calling and texting me with different numbers. Till date he has taken 10 sim cards. Torturing my family members every day with 40 different numbers. My life has become hell. I have lost my job. Sole bread winner of the family with 2 aged parents 70+ years. Nobody is able to help me. \\ \hline
She was out shopping at a supermarket when she noticed a 35-40 year old man was taking her pictures/videos. Initially, she thought that it might just be a misunderstanding and the man must just be using his phone. But later got too suspicious and scary because he was following her where ever she was going. She even reported the incident to the staff members of that supermarket but before any actions were taken the man had escaped. \\ \hline
I was harassed at my workplace in 2015 at ABC Technology Solutions while working as a Programmer Analyst Trainee by a senior employee in the team who attempted to establish physical contact/advances several times and I was unable to react and I later complained to my reporting manager. Upon complaining to the HR and my reporting managers they claimed that they know rules pertaining to the sexual harassment act and did not take any corrective/legal action and within a few days, I was asked to give forced resignation/termination. \\ \hline
This incident took place around 7 30 in the night. I took a bus home after my college trip and while I was waiting to collect the ticket from the conductor he touched me in my private part in the upper part of my body. I was too confused and scared to speak out something. After a while I got a seat in the bus and was continuously yelled at for sitting crossing my legs and was threatened to be thrown out of the bus. It was late at night and I just wanted to get home safely \\ 
\thickhline
\end{tabular}
\caption{Posts selected by \ALGO are more detailed and provide a clearer description of the incident. Detailed posts enable stakeholders to gain a deeper understanding of the incident's context, facilitating more informed and effective decision making.}
\label{tab:sample_posts}
\end{table*}

\end{document}